\documentclass[conference]{IEEEtran}

\ifCLASSINFOpdf
  \usepackage[pdftex]{graphicx}
\else
   \usepackage[dvips]{graphicx}
\fi
\usepackage{amsmath}
\usepackage{algorithm,algpseudocode}
\usepackage{array}
\usepackage{fixltx2e}
\usepackage{stfloats}
\usepackage{url}
\usepackage{multirow}
\providecommand{\keywords}[1]{\textbf{\small\textit{Keywords---#1}}}

\providecommand{\url}[1]{#1}
\csname url@samestyle\endcsname

\providecommand{\BIBentrySTDinterwordspacing}{\spaceskip=0pt\relax}
\providecommand{\BIBentryALTinterwordstretchfactor}{4}
\providecommand{\BIBentryALTinterwordspacing}{\spaceskip=\fontdimen2\font plus
\BIBentryALTinterwordstretchfactor\fontdimen3\font minus
  \fontdimen4\font\relax}
\providecommand{\BIBforeignlanguage}[2]{{%
\expandafter\ifx\csname l@#1\endcsname\relax
\typeout{** WARNING: IEEEtran.bst: No hyphenation pattern has been}%
\typeout{** loaded for the language `#1'. Using the pattern for}%
\typeout{** the default language instead.}%
\else
\language=\csname l@#1\endcsname
\fi
#2}}
\providecommand{\BIBdecl}{\relax}

\begin{document}
\title{Using Automatic Generation of Relaxation Constraints to Improve the Preimage Attack on 39-step MD4}

\author{\IEEEauthorblockN{Irina Gribanova, Alexander Semenov}
\IEEEauthorblockA{Matrosov Institute for System Dynamics and Control Theory SB RAS, Irkutsk, Russia\\
Email: the42dimension@gmail.com, biclop.rambler@yandex.ru
}
}

\maketitle

\begin{abstract}
In this paper we construct preimage attack on the truncated variant of the MD4 hash function. Specifically, we study the MD4-39 function defined by the first 39 steps of the MD4 algorithm. We suggest a new attack on MD4-39, which develops the ideas proposed by H. Dobbertin in 1998. Namely, the special relaxation constraints are introduced in order to simplify the equations corresponding to the problem of finding a preimage for an arbitrary MD4-39 hash value. The equations supplemented with the relaxation constraints are then reduced to the Boolean Satisfiability Problem (SAT) and solved using the state-of-the-art SAT solvers. We show that the effectiveness of a set of relaxation constraints can be evaluated using the black-box function of a special kind. Thus, we suggest automatic method of relaxation constraints generation by applying the black-box optimization to this function. The proposed method made it possible to find new relaxation constraints that contribute to a SAT-based preimage attack on MD4-39 which significantly outperforms the competition. 
\end{abstract}

\medskip 
\medskip 

\IEEEpeerreviewmaketitle

\section{Introduction}
Cryptographic hash functions have a wide range of applications: starting from various data security \cite{SchneierDBLP:books/daglib/0078909} and cryptocurrency protocols \cite{Nakamoto_bitcoin:a} to theoretical methods for cryptographic resistance justification of different cryptosystems \cite{Pointcheval2000, Koblitz:2015:ROM:2834659.2834706}. The Merkle-Damgard construction \cite{DBLP:conf/crypto/Merkle89,DBLP:conf/crypto/Damgard89a} is considered to be one of the most successful paradigms for constructing cryptographically resistant hash functions. The MD4 hash function \cite{DBLP:conf/crypto/Rivest90} is one of the first examples of hash functions based on the Merkle-Damgard construction. The widely known works \cite{DBLP:conf/eurocrypt/WangLFCY05, DBLP:conf/eurocrypt/WangY05} demonstrated the possibility of constructing collisions for hash functions MD4 and MD5. Thus, these functions have been compromised with respect to the \textit{collision attack}. However, today even MD4 remains resistant to the so called \textit{preimage attack}, which consists in the following: for a known hash value to find a corresponding input message. In this context, the implementation of preimage attacks on truncated variants of MD4 hash function is of interest. The truncated variant of the MD4 hash function is a variant of the original algorithm, which contains fewer steps (non-truncated variant consists of 48 steps). Hereinafter by MD4-k we denote a truncated variant of MD4 with $k$ steps, $k < 48$.

The first successful attack on truncated variant of MD4 with a relatively large number of steps was described by H. Dobbertin in \cite{Dobbertin:1998:FTR:647933.740752}. In this work it was showed that two-round version of MD4, i.e. MD4-32, is not one-way. The main idea of Dobbertin's attack is to use additional constraints on chaining variables at the certain steps of the MD4 algorithm to derive additional information, which leads to fast resolution of the corresponding cryptanalysis equations.

To the best of our knowledge, the attack described in \cite{DBLP:conf/SAT/DeKV07} is currently the best known attack on truncated variants of MD4. This attack is a SAT-variant of Dobbertin's attack which used the constraints of Dobbertin's type, in the sense that they were applied to the same chaining variables as in \cite{Dobbertin:1998:FTR:647933.740752}. The resulting system of cryptanalysis equations was reduced to the Boolean Satisfiability Problem (SAT) and then solved using the \textsc{minisat} \cite{DBLP:conf/sat/EenS03} SAT solver. For MD4-32 the SAT variant of the Dobbertin's attack turned out to be very effective. The main novelty of \cite{DBLP:conf/SAT/DeKV07} is to use Dobbertin's constraints and state-of-the-art SAT solvers to find preimages for MD4-k, $k = \{36,37,39\}$, within a reasonable time. However, it should be noted that the corresponding computational experiment for MD4-39 took a lot of time (about 8 hours on one processor core). In addition, in \cite{DBLP:conf/SAT/DeKV07} only the hash values of special kinds were considered. It is surprising that until 2017 there was, apparently, no progress in the practical implementation of the preimage attacks, which would be more effective than the attack from \cite{DBLP:conf/SAT/DeKV07}.

In \cite{PACT-2017} we presented a parallel SAT-variant of Dobbertin's attack on MD4-k, $k \leq 39$. 
One of the main results of \cite{PACT-2017} is the automatic search procedure of Dobbertin's constraints. For MD4-39 it was achieved a relatively fast solving of the preimage finding problem for the hash value $1^{128}$ (hereinafter $a^n$ denotes a word which consists of $n$ $a$ symbols).

In the present paper we improve the results from \cite{PACT-2017} in the following directions. First, we consider the problem of finding relaxation constraints of Dobbertin's type as a problem of black-box optimization over Boolean hypercube. To solve this problem we develop metaheuristic algorithm related to the class of Tabu Search algorithms. Using this algorithm we construct new relaxation constraints of Dobbertin's type for the MD4-39 preimage finding problem. These constraints, which are different from the ones presented in \cite{Dobbertin:1998:FTR:647933.740752} and \cite{DBLP:conf/SAT/DeKV07}, make it possible to find the MD4-39 preimages for 65-75\% of randomly generated 128-bit vectors within one minute of the \textsc{minisat2.2} SAT solver runtime on a single processor core Intel i7-3770K (3.5 GHz). Whereas using constraints from \cite{Dobbertin:1998:FTR:647933.740752} and \cite{DBLP:conf/SAT/DeKV07} \textsc{minisat2.2} is not capable to solve these tasks in several hours.

\section{Preliminaries}

As it was mentioned above, the MD4 hash function is a cryptographic hash function based on the Merkle-Damgard construction. This hash function can be used to calculate hash values for messages of an arbitrary length. The input message is split into 512-bit blocks. The resulting hash value is written in a special 128-bit register called \textit{hash register}. The hash register is divided into four parts of 32-bit length. According to the Merkle-Damgard construction, the fixed Initial Value (IV) is written to the hash register before hashing the first block of input message. Further, the contents of the hash register is iteratively modified. Thus, before hashing the 512-bit block with number $t+1$, the hash register contains the result of hashing of message blocks with numbers from 1 to $t$. The process of hashing of one 512-bit block is divided into 3 rounds with 16 steps each (thus, 48 steps in total). The contents of the hash register is mixed with the input message using the \textit{round functions}. In total, MD4 uses three round functions, detailed descriptions of which can be found in a variety of sources (i.e. \cite{DBLP:conf/eurocrypt/WangLFCY05}). On each step with number $k = 1,\ldots,48$ a variable called \textit{chaining variable} is associated with one of four parts of the hash register.

Hereinafter, we consider the problem of finding preimage (preimage attack) for the function of the kind:
\begin{equation}
\label{f_md4}
f_{MD4-k}: \{0,1\}^{512} \rightarrow \{0,1\}^{128},
\end{equation}
assuming that at the initial moment of time the hash register contains IV, corresponding to the specification of MD4. In fact, we consider the problem of finding 512-bit MD4-k preimage for known 128-bit hash value. Herein the main object of further interest is the function $f_{MD4-39}$.

Let us briefly recall the idea of the Dobbertin's attack \cite{Dobbertin:1998:FTR:647933.740752}. Based on the analysis of the round functions properties, H. Dobbertin proposed to fix with constant $K$ the values of certain chaining variables corresponding to the steps of the algorithm with numbers:
\begin{equation}
\label{set_1}
13, 14, 15, 17, 18, 19, 21, 22, 23, 25, 26, 27.
\end{equation}

The substitution of corresponding values into the cryptanalysis equations makes it possible to derive a significant part of the values of variables, which encode the unknown 512-bit input message. This, in turn, leads to a further simplification of the problem. As a result, in 1998 H. Dobbertin managed to find preimages for the MD4-32 hash function on a personal computer. We will refer to \textit{Dobbertin's constraints} to denote the additional constraints of the form $v_i = K$, where $v_i$ is chaining variable at $i$-th step and $i$ goes through the set of numbers from (\ref{set_1}). In general case, similar constraints on various steps of the MD4 algorithm different from (\ref{set_1}) can be used. For all such constraints we use the term \textit{relaxation constraints}.

The next step is to use a powerful combinatorial algorithm for solving the cryptanalysis equations with additional relaxation constraints. As it was mentioned above, this idea was proposed in \cite{DBLP:conf/SAT/DeKV07} where Dobbertin's constraints were used with constant $K = 0$ and the corresponding cryptanalysis equations were solved using the \textsc{minisat} \cite{DBLP:conf/sat/EenS03} SAT solver.

Let us recall, that SAT (short for "Satisfiability") is a problem of satisfiability of an arbitrary Boolean formula, which consists in the following: for an arbitrary formula $F$ over the set of Boolean variables $X$ to decide if there exists such an assignment of variables from $X$ that makes this formula true. It is usually considered in the variant where $F$ is presented in conjunctive normal form (CNF).

The approach in which modern SAT solvers are used to solve cryptanalysis problems is called \textit{SAT-based cryptanalysis}. To reduce the preimage finding problem (inversion problem) of an arbitrary total discrete function of the kind $f: \{0,1\}^n \rightarrow \{0,1\}^m$ to SAT one can use various automatic translation systems, like \textsc{Cryptol} \cite{DBLP:conf/plpv/ErkokM09} or \textsc{URSA} \cite{journals/lmcs/predrag}. In our work we use software system \textsc{Transalg} \cite{DBLP:conf/ecai/OtpuschennikovS16} specially designed to produce SAT encodings for the inversion problems of cryptographic functions. \textsc{Transalg} performs a symbolic execution \cite{DBLP:journals/cacm/King76} of a program, which specifies the considered function $f$. The result of such execution is a CNF $C(f)$ called \textit{template CNF}. By $C(f, y)$ we denote the result of the substitution of a known image $y$ of function $f$, $y \in Range f$, into CNF $C(f)$. It can be shown that $C(f, y)$ is satisfiable. Asumming that satisfying assignment for $C(f, y)$ is found using SAT solver, a preimage $x \in \{0,1\}^n$ such that $f(x) = y$ can be extracted from this assignment. Using the methods of SAT-based cryptanalysis to find preimages of cryptographic hash functions is called a \textit{SAT-based preimage attack} on this function.

As it was mentioned above, in \cite{PACT-2017} the parallel version of SAT-based preimage attack on MD4-k ($k \leq 39$) from \cite{DBLP:conf/SAT/DeKV07} was proposed. However, in the role of relaxation constraints the same Dobbertin's constraints were used. In the next section we consider the generation of relaxation constrains as a problem of block-box optimization over Boolean hypercube. We also present computational results obtained using new relaxation constraints.

\section{The generation of relaxation constrains as a problem of black-box optimization}
Let us consider the preimage finding problem for the function of the kind (\ref{f_md4}) with fixed $k$ and reduce this problem to SAT. Let $C(f_{MD4-k} )$ be template CNF for this problem and $X$ be a set of all Boolean variables in $C(f_{MD4-k} )$. By $C(f_{MD4-k}, \chi)$ we denote a CNF obtained by substitution of a hash value $\chi \in \{0,1\}^{128}$ into $C(f_{MD4-k})$.

Below we briefly describe the idea of switching variables introduced in \cite{PACT-2017}. Suppose that there is some set of relaxation constraints $\Re = \{R_1 , \ldots , R_Q\}$, where an arbitrary constraint $R_j, j \in \{1, \ldots , Q\}$ is usually a conjunction of some literals, i.e. a formula of the kind:
\begin{equation}
\label{form_l}
l_1(x_{j_1})\wedge \ldots \wedge l_{t_j}(x_{j_{t_j}}), \{x_{j_1}, \ldots, x_{j_{t_j}}\} \subseteq X
\end{equation}
(note that literal $l(x)$ is a formula of the kind $x$ or $\lnot x$, where $x$ is a Boolean variable).

Consider a new set of Boolean variables $S=\{s_1, \ldots , s_Q \}$, $S \cap X = \emptyset$. We call such variables \textit{switching variables}. Let us associate with an arbitrary $R_j, j \in \{1, \ldots , Q\}$ of the kind (3) the following CNF:
$$
C_{R_j} = (\lnot s_j \vee l_1(x_{j_1})) \wedge \ldots \wedge (\lnot s_j \vee l_{t_j}(x_{j_{t_j}})).
$$
It should be noted that the literals $l_1(x_{j_1}), \ldots , l_{t_j} (x_{j_{t_j}})$ can be derived from CNF $s_j \wedge C_{R_j}$ using the Unit Propagation (UP) rule \cite{Dowling1984}. Then, this new information will be further propagated according to UP. On the other hand, it's obvious that $\lnot s_j \wedge C_{R_j} \equiv \lnot s_j$. In this case the constraint $R_j$ does not give any additional information. We say that the constraint $R_j$ is \textit{active} if $s_j = 1$ and \textit{inactive} if $s_j = 0$.

Let $S$ be the set of switching variables. The set of all possible values of variables from $S$ is $\{0,1\}^Q$. Thus, each nonzero Boolean vector $\lambda \in \{0,1\}^Q$ specifies some set of active relaxation constraints from set $\Re$. Our first goal is to learn how to distinguish more effective sets of relaxation constraints from less effective ones (in the sence of increasing the efficiency of the corresponding SAT-based preimage attack). To solve this problem we use the approach similar to that applied in \cite{10.1007/978-3-319-21909-7_21,Semenov2016} for searching SAT partitionings \cite{DBLP:phd/basesearch/Hyvarinen11} of SAT-instances arising in cryptanalysis problems. In particular, we introduce a measure of efficiency for an arbitrary set of relaxation constraints from $\Re$ and consider the problem of finding sets of relaxation constrains with good efficiency as a problem of maximization of a specially defined function over Boolean hypercube $\{0,1\}^Q$.

The problem of choosing the adequate measure of efficiency for relaxation constraints is quite non-trivial. At this stage, after a large number of experiments, the measure was defined as follows.
Consider an arbitrary vector $\lambda \in \{0,1\}^Q$, where $\{\lambda_{h_1} , \ldots , \lambda_{h_d} \}$ is a set of components equal to 1. Taking into account the above, these components define a set of active relaxation constraints from $\Re = \{R_1 , \ldots , R_Q\}$, namely, the constraints with numbers $h_1 , \ldots , h_d$. Consider the following CNF:
\begin{equation}
\label{C_tilda}
\tilde{C}(\lambda) = C(f_{MD4-k}, \chi) \wedge (\wedge_{j\in \{h_1,\ldots,h_d\}}C_{R_j}).
\end{equation}

Everywhere below, we will use notation $\tilde{C}(\lambda) \rightarrow_{UP} l(x)$ to denote that literal $l(x)$ is derived from CNF (\ref{C_tilda}) using UP. By $X^{in} \subset X$ we denote a set of Boolean variables in $C(f_{MD4-k} ,\chi)$, which encode an unknown 512-bit input of $f_{MD4-k}$ function.

For an arbitrary $\lambda \in \{0,1\}^Q$ we consider the function:
\begin{equation}
\label{m_function}
\mu(\lambda) = \#\{l(x)|\tilde{C}(\lambda) \rightarrow_{UP} l(x):x \in X^{in}\}.
\end{equation}
In other words $\mu(\lambda)$ is the number of literals from $X^{in}$, which were derived by UP from CNF (\ref{C_tilda}) as a result of activation of relaxation constraints corresponding to vector $\lambda$.

We will consider the maximization problem of (\ref{m_function}) over Boolean hypercube $\{0,1\}^Q$. It's obvious that function (\ref{m_function}) is a function of black-box type and its analitical properties are unknown. Thereby it is justified to use metaheuristic algorithms for the maximization problem of (\ref{m_function}). At this stage, we implemented a special variant of Tabu Search algorithm \cite{DBLP:books/daglib/0093574}. In the computational experiments, discussed further, we considered the Hamming neighborhoods of the radius 1 in $\{0,1\}^Q$. The pseudocode of the algorithm is presented below.

\begin{algorithm}
 \caption{Tabu Search Algorithm (A1)}
 \label{alg_a1}
 \begin{algorithmic}[1]
 \State $\langle \lambda_{center}, \mu_{best} \rangle \leftarrow \langle \lambda_{start}, \mu(\lambda_{start}) \rangle$\; 
 \State initializeLists($L_1, L_2$)
 \Repeat 
  \State bestValueUpdated $\leftarrow$ false
  \Repeat 
  	\State $\lambda \leftarrow$ getNewPoint($N(\lambda_{center}))$
  	\State markPointInTabuLists($\lambda,L_1, L_2$)
  	\If{isCorrectPoint($\lambda$)}
  	\State compute $\mu(\lambda)$
		 \If{$\mu(\lambda) > \mu_{best}$}
	   		\State $\langle \lambda_{best}, \mu_{best} \rangle \leftarrow \langle\lambda,\mu(\lambda) \rangle$
	   		\State bestValueUpdated $\leftarrow$ true
	   		%}
	   	\EndIf
   	\EndIf
  \Until $N(\lambda_{center})$ is checked
  \If{bestValueUpdated}
  \State $\lambda_{center} \leftarrow \lambda_{best}$\
  \Else
  \State $\lambda_{center} \leftarrow$ getNewCenter($L_2$)
  \EndIf
  \Until timeExceeded() or $L_2 = \emptyset $
  \State \textbf{return} $\langle \lambda_{best}, \mu_{best} \rangle$
 \end{algorithmic}
\end{algorithm}

Let us give more detailed description of the A1 algorithm. The input of A1 algorithm is CNF $C(f _{MD4-39}, \chi)$ encoding the MD4-39 preimage finding problem for a known hash value χ and starting point $\lambda_{start}$ with a corresponding set of relaxation constraints of the kind (\ref{form_l}). As a starting point, either a random point or some known point can be chosen. The contents of the $L_1$ and $L_2$ lists are initialized using function \textit{initializeLists}. At the initial moment the $L_1$ list is empty, $L_2$ contains point $\lambda_{start}$, $\lambda_{center}$ is equal to $\lambda_{start}$ and $\mu_{best}$ is the value of the objective function $\mu(\lambda_{start})$.

In the main loop of the algorithm the neighborhood of the point $\lambda_{center}$, denoted by $N(\lambda_{center})$, is considered. Function \textit{getNewPoint} chooses any unchecked point from $N(\lambda_{center})$ as a current point $\lambda$. Function \textit{markPointInTabuLists} adds point $\lambda$ to $L_2$ and then marks $\lambda$ as checked in all neighborhoods of points from $L_2$ which contain $\lambda$. This allows to avoid re-processing of the same points. If the neighborhood of some point contains only checked points, then this point is moved to $L_1$.

For current point $\lambda$ and corresponding CNF $\tilde{C}(\lambda)$ of the kind (\ref{C_tilda}) function \textit{isCorrectPoint} runs a SAT solver for a short period of time. If, as the result, CNF $\tilde{C}(\lambda)$ is proven to be unsatisfiable, then the algorithm moves to the next point from the neighborhood $N(\lambda_{center})$. Otherwise, the value $\mu(\lambda)$ is computed and compared with the value $\mu_{best}$.

In case if we did not improve $\mu_{best}$ value in the neighborhood of $\lambda_{center}$, new point $\lambda_{center}$ must be selected from $L_2$. Function \textit{getNewCenter} chooses a point from $L_2$ with a value of the objective function which is closest to the known $\mu_{best}$.

The algorithm is completed if a certain time limit is exceeded or the entire search space is processed (in this case $L_2$ is empty). The output of the algorithm is the point $\lambda_{best}$ and the corresponding value of the objective function $\mu(\lambda_{best})$.

\section{Computational experiments}

In this section we describe computational results for MD4-39 preimage attack using the method of relaxation constraints generation described above. At the current stage, the A1 algorithm is implemented as a single-threaded application. To calculate the value of the function (\ref{m_function}) the Unit Propagation procedure, implemented in all modern CDCL solvers, is used. 

Everywhere below, the constraints of the kind (\ref{form_l}), consisting only of literals with negation, were used as relaxation constraints. Thus, we used constraints of Dobbertin's type with constant $K = 0$.

Let us note here that the structure of the MD4-39 hash function makes it impossible to impose constraints on the first four and the last (preceding the calculation of the final hash value) four steps of the MD4-39 algorithm. According to this, the sets of new relaxation constraints were selected (using the values of the corresponding switching variables) from the set $\Re$ of power $Q = 31$. Thus, the problem of maximization of the function (\ref{m_function}) over Boolean hypercube $\{0,1\}^{31}$ was considered.

In the early experiments it was found that some sets of relaxation constraints produce CNFs, for which the UNSAT can be proven quite quickly (within a few seconds). In practical implementation of the algorithm for each set of values of switching variables that specifies a set of relaxation constraints, not only the value of function (5) was calculated, but also short time limit was given to solve the corresponding SAT instance.
This step allows to screen out some points without the computation of the objective function. 

In the A1 algorithm the following actions are performed: selection of starting search point; screening out the points for which unsatisfiability is proven quickly; accumulation of all record points; exit from local maxima.

The A1 algorithm was run on one core of Intel i7-3770K (3.5 GHz) processor under Linux OS (Ubuntu 16.04). In all computational experiments the MD4-39 preimage finding problem for $\chi = 0^{128}$ was considered. 
For the points, obtained using the A1 algorithm, with the value of the objective function close to the maximal possible value (i.e. 512), we established that corresponding sets of relaxation constraints define unsatisfiable CNFs.
The satisfiability problems of such CNFs were considered as a separate problems, which in some cases required a significant amount of time. Thus, it was necessary to select points from $\{0,1\}^{31}$ for which there was a good chance for the corresponding CNF of the kind (\ref{C_tilda}) to be satisfiable. 
To find such points the following heuristic was used: first, to select only those points where the value of the function $\mu(\lambda)$ was improved (i.e., record points); second, to select the points with the value of the function $\mu(\lambda)$ from the interval $[256,320]$. The total number of record points from the number of all points processed in several hours was approximately 2\%. The total number of perspective points identified by the heuristic described above was 0.5\%. For each point from the perspective set of points the \textsc{minisat2.2} SAT solver was applied to the corresponding CNFs with a small time limit (60 seconds).

As a result of the above actions, two new sets of relaxation constraints were obtained. 
These sets are specified by the following vectors of values of switching variables from $\{0,1\}^{31}$:
\[
\begin{array}{lc}
\rho_1: &0000000001101110111011101000000\\
\rho_2: &0000000000101110111011101100000\\
\end{array}
\]

The application of these sets allows one to find preimages of the MD4-39 hash function for known hash values $0^{128}$ and $1^{128}$ within a minute of \textsc{minisat2.2} runtime (whereas using constraints from \cite{DBLP:conf/SAT/DeKV07} the solution of the preimage finding problem for $1^{128}$ requires about 2 hours, and the preimage finding problem for $0^{128}$ cannot be solved in 8 hours). Corresponding results are presented in Table \ref{table_1}, where $\rho_{De}$ denotes the set of relaxation constraints described in \cite{DBLP:conf/SAT/DeKV07} and $\rho_{Dobbertin}$ denotes the variant of Dobbertin's constraints from \cite{Dobbertin:1998:FTR:647933.740752} with constant $K = 0$. Below these relaxation constraints are specified by the vectors of values of switching variables from $\{0,1\}^{31}$ (in the similar notation to that of $\rho_1$ and $\rho_2$):
\[
\begin{array}{lc}
\rho_{Dobbertin}: &0000000011101110111011100000000\\
\rho_{De}: &0000000001101110111011100000000\\
\end{array}
\]	

What is particularly interesting is that the application of new sets of relaxation constraints $\rho_1$ and $\rho_2$ also allows one to find preimages of MD4-39 for randomly generated 128-bit hash values persistently. To obtain this result, we considered a test set consisting of 500 randomly generated vectors from $\{0,1\}^{128}$. Each vector from this set was taken as a hash value of the MD4-39 hash function. After that the preimage finding problem for this value was solved using constraints $\rho_1$ and $\rho_2$. For the prevailing part of the tasks (65-75\%) the solutions were successfully found using the \textsc{minisat2.2} SAT solver. The average time of finding one preimage was less than 1 minute. The rest ones (25-35\% of the tasks) corresponded to 128-bit vectors for which there were no MD4-39 preimages under constraints $\rho_1$ and $\rho_2$. These results are presented in Table \ref{table_2}. Note that even in a few hours we did not manage to solve the preimage finding problem for any vector from the test set using constraints from \cite{Dobbertin:1998:FTR:647933.740752} or \cite{DBLP:conf/SAT/DeKV07}.

\section{Conclusion}
In the present paper a new SAT-based preimage attack on the 39-step variant of the MD4 cryptographic hash function is suggested. This attack makes it possible to solve the MD4-39 preimage finding problem for a very significant percentage of randomly generated 128-bit vectors, spending on one such vector less than a minute of \textsc{minisat2.2} runtime. The proposed attack is much more effective than the best known attack on the considered truncated variant of the MD4 hash function presented about 10 years ago in \cite{DBLP:conf/SAT/DeKV07}.

We intend to develop the approach described in this paper in the direction of studying the preimage finding problem of MD4-k, where $k \geq 40$. The preliminary results show that the corresponding problem for MD4-40 demands significantly more computational resources in comparison with MD4-39: the relaxation constrains constructed using the method described in this paper do not make it possible to solve the  MD4-40 preimage finding problem on a single processor core. At the same time, this effect is not observed between the preimage attacks on MD4-38 and MD4-39. In the nearest future we plan to apply the parallel SAT solvers to the inversion problems of MD4-k, $k \geq 40$ with relaxation constraints constructed using the method presented in this paper.

\begin{table}
\caption{Finding the MD4-39 preimages for hash values $0^{128}$ and $1^{128}$}
\label{table_1}
\centering
\begin{tabular}{|p{1.5cm}|p{1.5cm}|p{1.5cm}|p{1.5cm}|}
\hline
Relaxation & \multirow{2}{*}{$\mu(\lambda)$} & \multicolumn{2}{l|}{Solving time (s)} \\
\cline{3-4}
constraints &  &\multirow{2}{*}{$\chi = 0^{128}$}& \multirow{2}{*}{$\chi = 1^{128}$}\\  
& & & \\
\hline
$\rho_1$ & 288 & 20 & 10 \\
\hline
$\rho_2$ & 288 & 60 & UNSAT \\
\hline
$\rho_{Dobbertin}$ & 288 & 20 & $>$ 8 hours \\
\hline
$\rho_{De}$ & 256 &$>$ 8 hours & 7000 \\
\hline
\end{tabular}
\end{table}

\begin{table}
\caption{Finding the MD4-39 preimages for 500 randomly generated 128-bit hash values}
\label{table_2}
\centering
\begin{tabular}{|l|l|l|p{1.8cm}|p{1.7cm}|}
\hline
Relaxation& Avg.& Max.& \multicolumn{2}{l|}{Solved instances (in \% of total} \\ 
constraints &solving&solving& \multicolumn{2}{l|}{number of instances)} \\
\cline{4-5}
&time (s)&time (s)&with preimages (SAT)&with no preimages (UNSAT)\\
\hline
$\rho_1$& 12 & 80 & 65 & 35 \\ \hline
$\rho_2$& 46 & 250 & 75 & 25 \\ \hline

\end{tabular}
\end{table}

\section{Related Work}
The first mention of the approach to the construction of hash functions, which is widely known today as the Merkle-Damgard construction, can be found in \cite{Merkle_PHD}. In \cite{DBLP:conf/crypto/Merkle89} and \cite{DBLP:conf/crypto/Damgard89a} R. Merkle and I. Damgard independently described a number of important properties of hash functions based on this construction. One of the first practical implementations of Merkle-Damgard construction was the MD4 hash function \cite{DBLP:conf/crypto/Rivest90} developed by R.Rivest. In \cite{DBLP:conf/eurocrypt/WangLFCY05} the MD4 hash function was completely compromised with respect to the collision attack. The collision search problem for the functions from MD family in the form of SAT was first proposed in \cite{DBLP:conf/frocos/JovanovicJ05}. However, real practical results in this direction were obtained later in \cite{DBLP:conf/SAT/MironovZ06}. The use of propositional encodings presented in \cite{DBLP:conf/ecai/OtpuschennikovS16} made it possible to find collisions for MD4 hash function (with the help of modern SAT solvers) about 1000 times faster than it was done in \cite{DBLP:conf/SAT/MironovZ06}.

In a number of works the resistance of MD4 hash function to the preimage attack was studied. Today it is generally accepted that MD4 is not resistant to the preimage attack, although the best known preimage attack on the full-round version of MD4 is theoretical \cite{Leurent:10.1007/978-3-540-71039-4_26}. The first practical preimage attack on truncated variant of MD4 was implemented by H.Dobbertin: the algorithm presented in \cite{Dobbertin:1998:FTR:647933.740752} allows one to find preimages of MD4-32 on a personal computer. As far as we know, in the last 10 years the best practical attack on truncated variants of MD4 was attack described in \cite{DBLP:conf/SAT/DeKV07}. In this attack the \textsc{minisat} \cite{DBLP:conf/sat/EenS03} SAT solver was used to find preimages of MD4-39 weakened by the additional constraints. In the present paper we significantly improve the results presented in \cite{DBLP:conf/SAT/DeKV07}.

% use section* for acknowledgment
\section*{Acknowledgment}
The research was funded by Russian Science Foundation (project No. 16-11-10046).

\end{document}